\documentclass{bmvc2k}

\usepackage{times}
\usepackage{epsfig}
\usepackage{graphicx}
\usepackage{amsmath}
\usepackage{amssymb}
\usepackage{mwe}
\usepackage{multirow}
 
\usepackage[export]{adjustbox}
\usepackage{float}
\usepackage{dashrule}
\usepackage{arydshln}

\usepackage{booktabs,colortbl}

\usepackage[utf8x]{inputenc}

\title{Reducing Label Noise in Anchor-Free Object Detection}

\addauthor{Nermin Samet}{nermin@ceng.metu.edu.tr}{1}
\addauthor{Samet Hicsonmez}{samethicsonmez@hacettepe.edu.tr}{2}
\addauthor{Emre Akbas}{emre@ceng.metu.edu.tr}{1}

\addinstitution{
Middle East Technical University\\
Ankara, Turkey
}
\addinstitution{
Hacettepe University\\
Ankara, Turkey\\
}

\runninghead{Samet, Hicsonmez, Akbas}{PPDet}

\begin{document}

\maketitle

\begin{abstract}

Current anchor-free object detectors label all the features that spatially fall inside a predefined central region of a ground-truth box as positive. This approach causes label noise during training, since some of these positively labeled features may be on the background or an occluder object, or they are simply not discriminative features. In this paper, we propose a new labeling strategy aimed to reduce the label noise in anchor-free detectors. We sum-pool predictions stemming from individual features into a single prediction. This allows the model to reduce the contributions of non-discriminatory features during training. We develop a new one-stage, anchor-free object detector, PPDet, to employ this labeling strategy during training and a similar prediction pooling method during inference. On the COCO dataset, PPDet achieves the best performance among anchor-free top-down detectors and performs on-par with the other state-of-the-art methods. It also outperforms all major one-stage and two-stage methods in small object detection (${AP}_{S}$ $31.4$). Code is available at \url{https://github.com/nerminsamet/ppdet}.

\end{abstract}


\section{Introduction}
\label{sec:intro}

Early deep learning based object detectors were two-stage, proposal driven methods~\cite{faster, fastrcnn}. In the first stage, a sparse set of object proposals are generated and a convolutional neural network (CNN) categorizes them in the second stage. Later, the idea of unified detection in a single stage has gained increasing attention~\cite{ssd,yolo3,retinanet, dssd}, where proposals were replaced with predefined anchors. On the one hand, anchors have to cover the image densely (in terms of location, shape and scale) so as to maximize recall; on the other hand, their number should be kept at a minimum to reduce both the inference time and the imbalance problems \cite{imbalance} they create during training.

A considerable amount of effort has been spent on addressing the drawbacks of anchors: several methods have been proposed to improve the quality of anchors \cite{region_proposals, metanachor}, to address the extreme foreground-background imbalance \cite{ohem, retinanet, imbalance}, and recently, one-stage anchor-free methods have been developed.  There are two main groups of prominent approaches in anchor-free object detection. The first group is keypoint based, bottom-up methods, popularized after the pioneering work CornerNet~\cite{cornernet}. These detectors \cite{cornernet, extremenet, centernet, triplet} first detect keypoints (e.g. corners, center and extreme points) of objects, and then group them to yield whole-object detections. The second group of anchor-free object detectors \cite{fcos, foveabox, fsaf} follow a top-down approach, and directly predict class and bounding box coordinates at each location in the final feature map(s).

\begin{figure}
\centering
\begin{tabular}{cccc}
\includegraphics[width=0.28\textwidth]{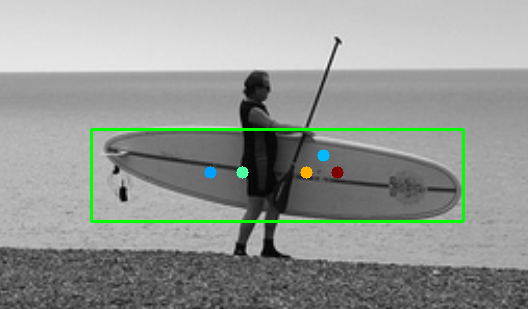} &
\includegraphics[width=0.28\textwidth]{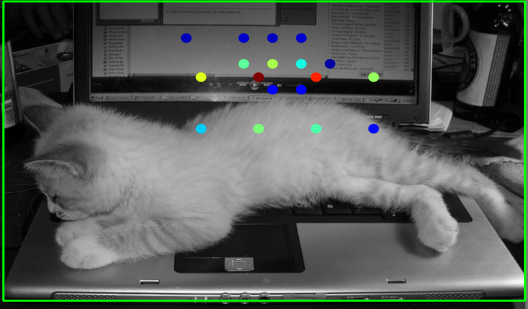}&
\includegraphics[width=0.28\textwidth]{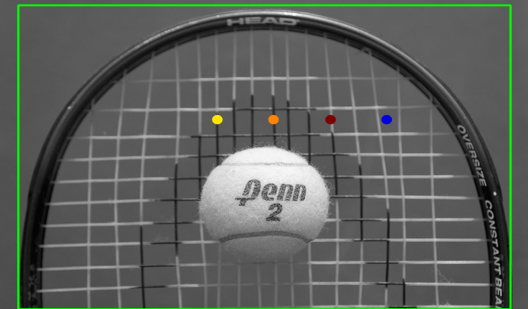}&
\includegraphics[width=0.03\textheight]{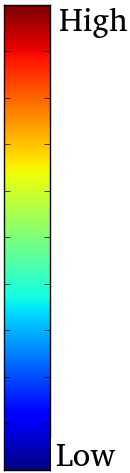} \\
\end{tabular}
\caption{Three sample detections by PPDet, from left to right: surfboard, laptop and racket. The colored dots show the locations whose predictions are pooled to generate the final detection shown in the green bounding box. The color denotes the contribution weight. Highest contributions are coming from the objects and not occluders or background areas. Images are from COCO \texttt{val2017} set.}
\label{fig:teaser}
\end{figure}

One important aspect of object detector training is the strategy used to label object candidates, which could be proposals, anchors or locations (i.e. features) in the final feature map. In order to label a candidate ‘positive’ (foreground) or ‘negative’ (background) during training, a variety of  strategies have been proposed, based on \textit{Intersection over Union (IoU)}~\cite{faster, rfcn, ssd, retinanet}, \textit{keypoints}~\cite{cornernet, extremenet, centernet, triplet} and \textit{relative location to a ground-truth box}~\cite{fcos, guided, foveabox}. Specifically in top-down anchor-free object detectors, after the input image is passed through the backbone feature extractor and the FPN \cite{fpn}, features that spatially fall inside a ground-truth box are labeled as positive and others as negative -- there is also an “ignore” region in between. Each of these positively-labeled features contributes to the loss function as a separate prediction. The problem with this approach is that some of these positive labels might be plain-wrong or of poor quality, hence, they inject label noise during training. Noisy labels come from (i) non-discriminatory features that are on the object, (ii) background features within the ground-truth box, and (iii) occluders (Fig.~\ref{fig:teaser}). In this paper, we propose an anchor-free object detection method, which relaxes the positive labeling strategy so that the model is able to reduce the contributions of non-discriminatory features during training. In accordance with this training strategy, our object detector  employs  an inference method where highly-overlapping predictions enforce each other. 
        
In our method, during training, we define a ``positive area’’ within a ground-truth (GT) box, which is co-centric and has the same shape with the GT box. We experimentally adjust the size of the positive area relative to the GT box. As this is an anchor-free method, each feature (i.e. location in the final feature maps) predicts a class probability vector and bounding box coordinates. The class predictions from the positive area of a GT box get pooled together and contribute to the loss as a single prediction. This sum-pooling alleviates the noisy-labels problem mentioned above since the contributions of features from non-object (background or occluded) areas, and non-discriminatory features are automatically down weighted during training.  
At inference, class probabilities of highly overlapping boxes are again pooled together to obtain the final class probabilities. We name our method as “PPDet”, which is short for ``prediction pooling detector.’’

Our contributions with this work are two fold: (i) a relaxed labelling strategy, which allows the model to reduce the contribution of non-discriminatory features during training, and (ii) a new object detection method, PPDet, which uses this strategy for training and a new inference procedure based on prediction pooling.  We show the effectiveness of our proposal on the COCO dataset. PPDet outperforms all anchor-free top-down detectors and performs on-par with the other state-of-the-art methods. PPDet is especially effective for detecting small objects ($31.4$ \textit{AP$_{S}$}, better than state-of-the-art).

\section{Related Work}

Apart from the classical one-stage~\cite{ssd, yolo3, retinanet, dssd} vs. two-stage~\cite{ faster, fastrcnn, rfcn} categorization of object detectors, we can also categorize the current approaches into two: anchor-based and anchor-free. Top-down anchor-free object detectors simplify the training process by eliminating complex IoU operations and focus on identifying the regions that may contain objects. In that sense, FCOS~\cite{fcos}, FSAF~\cite{fsaf} and  FoveaBox~\cite{foveabox} first map GT boxes onto the FPN levels, then label the locations, i.e. features, as positive or negative based on whether they are inside a  GT box. Bounding box prediction is only for positively-labeled locations. FoveaBox~\cite{foveabox} and  FSAF~\cite{fsaf} define three areas for each object instance; \textit{positive} area, \textit{ignore} area and \textit{negative} area.  FoveaBox defines the \textit{positive} (\textit{fovea}) area as the region which is co-centric with the GT box, and whose dimensions are scaled by  a  (shrink) factor $0.3$. All locations within this positive area are labeled as positive. Similarly, another  area is obtained using a  shrink factor of $0.4$. Any location that is outside this area is  labeled  as negative. If a location  is neither positive nor negative, it is ignored during training.  FSAF follows the same approach and uses shrink factors $0.2$ and $0.5$, respectively. 
Instead of having pre-defined discrete areas as in~\cite{fsaf, foveabox, guided}, FCOS down-weights the features based on their distance  to the center using a centerness branch. FCOS and FoveaBox implement static feature-pyramid level selection where they assign objects to levels based on GT box scale and GT box regression distance, respectively. Unlike them, FSAF relaxes the feature selection step and dynamically assigns each object to the most suitable feature-pyramid level. 
  
Bottom-up anchor-free object detection methods \cite{cornernet, extremenet, centernet, triplet} aim to detect certain keypoints of objects, such as corners and the center. Their labeling strategy uses heatmaps, and in this sense, it is considerably different from that of top-down anchor-free methods. More recently, HoughNet, a novel, bottom-up voting-based method that can utilize both near and long-range evidence to detect object centers, has shown comparable performance with major one-stage and two-stage top-down methods \cite{houghnet}. 

In the anchor-based approaches~\cite{faster, rfcn, ssd, yolo3, retinanet, freeanchor, guided}, objects are predicted from regressed anchor boxes. During training, the label of an anchor box is determined  based on its  intersection over union (IoU) with a GT box. Different detectors use different criteria, e.g. Faster RCNN~\cite{faster} labels an anchor as positive if $IoU>0.7$, and negative if $IoU<0.3$; R-FCN~\cite{rfcn}, SSD~\cite{ssd} and Retinanet~\cite{retinanet} use $IoU>0.5$ for positive labeling but slightly different criterias for negative labeling. There are two prominent anchor-based methods which directly address the labeling problem.  Guided Anchoring~\cite{guided} introduces a new adaptive anchoring scheme that  learns arbitrary shaped boxes instead of dense and predefined ones. Similar to FSAF~\cite{fsaf}, FoveaBox \cite{foveabox} and our method PPDet,  Guided Anchoring follows region based labelling and defines three types of regions for each ground-truth object; \textit{center} region, \textit{ignore} region and \textit{outside} region, and labels the generated anchors positive if it resides inside the \textit{center} region, negative if in \textit{outside} region and ignores the rest. On the other hand, FreeAnchor~\cite{freeanchor} applies the idea of relaxing positive labels for anchor-based detectors. This is the most similar method to ours.  It replaces hand-crafted anchor assignment with a maximum likelihood estimation procedure, where anchors are set free to choose their  GT box. Since FreeAnchor is optimizing object-anchor matching using a customized loss function, it can not be directly applied to anchor-free object detectors.


\begin{figure}
\centering
      \includegraphics[width=0.9\linewidth]{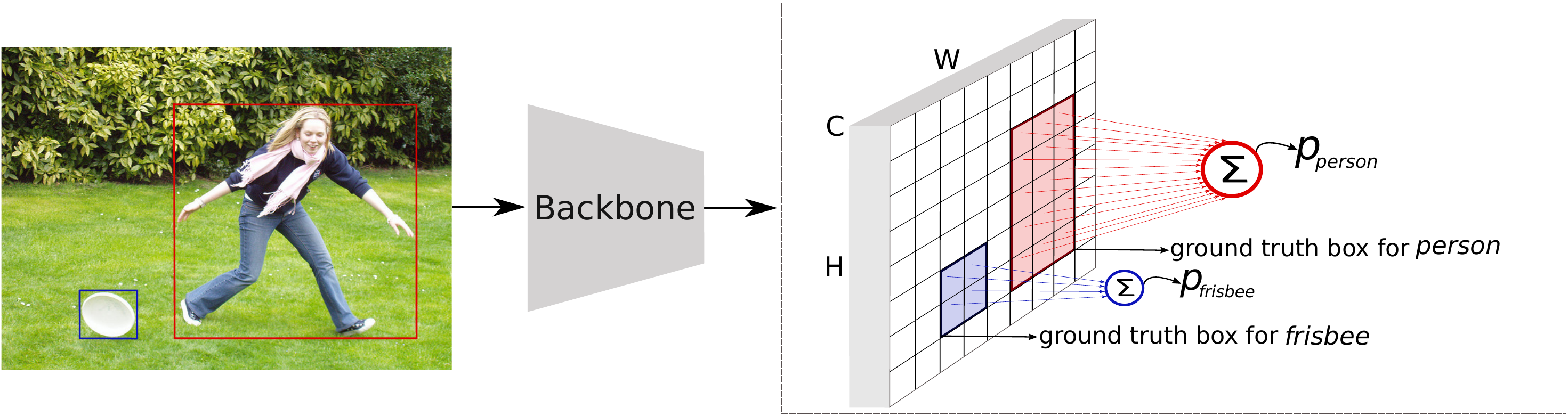} 
  \caption{Prediction pooling during training of  \textit{PPDet}. For simplicity, it is illustrated on a single FPN level and the bounding box regression branch is not shown. Blue and red cells are foreground cells. Same color foreground cells, each of which is a $C$-dimensional vector, are pooled, i.e. summed  together, to form the final prediction score for the corresponding object. These pooled scores (i.e., $p_\mathrm{person}, p_\mathrm{frisbee}$), are fed to the loss function (i.e., focal loss).(Best viewed in color.)
}
\label{fig:modelv1} 
\end{figure}
\section{ Methods}

\paragraph{Labeling strategy and training.} Anchor-free detectors limit prediction of GT boxes by assigning them to appropriate FPN levels based on their  scales~\cite{foveabox} or target regression distances~\cite{fcos}. Here, we follow the scale-based assignment strategy~\cite{foveabox} since it is a way of  naturally associating GT boxes with feature pyramid levels. Then, we construct two different regions for each GT box. We define the ``positive area’’ as the region that is co-centric with the GT box and having the same shape as the GT box. We experimentally set  the size of the ``positive area’’.  Then, we identify all the locations (i.e. features)  that spatially fall inside the ``positive area’’ of a GT box as ``positive (foreground)’’ features  and the rest as ``negative (background)’’ features. Each positive feature  is assigned to the ground-truth box that contains it. In Figure~\ref{fig:modelv1}, blue and red cells represent foreground cells and the rest (empty or white) are background cells. The blue cells are assigned to the \textit{frisbee} object and the red cells to the \textit{person} object. To obtain the final detection score for an object instance, we pool the classification scores of all the features that are assigned to that object, by adding them together to obtain a final $C$-dimensional vector where $C$ is the number of the classes. All features except the positively labelled ones are negatives. Each negative feature contributes individually to the loss (i.e. no pooling). This final prediction vector is fed to the focal loss (FL). For example, suppose $\{\mathbf{p}_i | i=1,2,\dots,N\}$ represent the red, foreground features that are assigned to the person object in Figure~\ref{fig:modelv1}. Let $\mathbf{y}$ be the ground-truth, one-hot vector for the person class. Then, this particular object instance contributes ``FL$(\sum_i \mathbf{p}_i, \mathbf{y})$’’ to the loss function in training. Each object instance is represented with a single prediction.

By default, we assign positive features  to the object instance of the box they are in. At this point, assignment of features  in the intersection areas of different GT boxes is an issue to be handled. In such cases, we assign those features to the GT box with the smallest distance to their centers.
Similar to other anchor-free methods~\cite{fcos, fsaf, foveabox, centernet}, in our model each foreground feature assigned to an object is trained to predict the coordinates of its object’s GT  box.

We use the focal loss~\cite{retinanet} ($\alpha=0.4$ and $\gamma=1.5$) for the classification branch and smooth ${\ L_{1}}$ loss~\cite{fastrcnn} for the regression branch.


\begin{figure}
      \includegraphics[width=\linewidth]{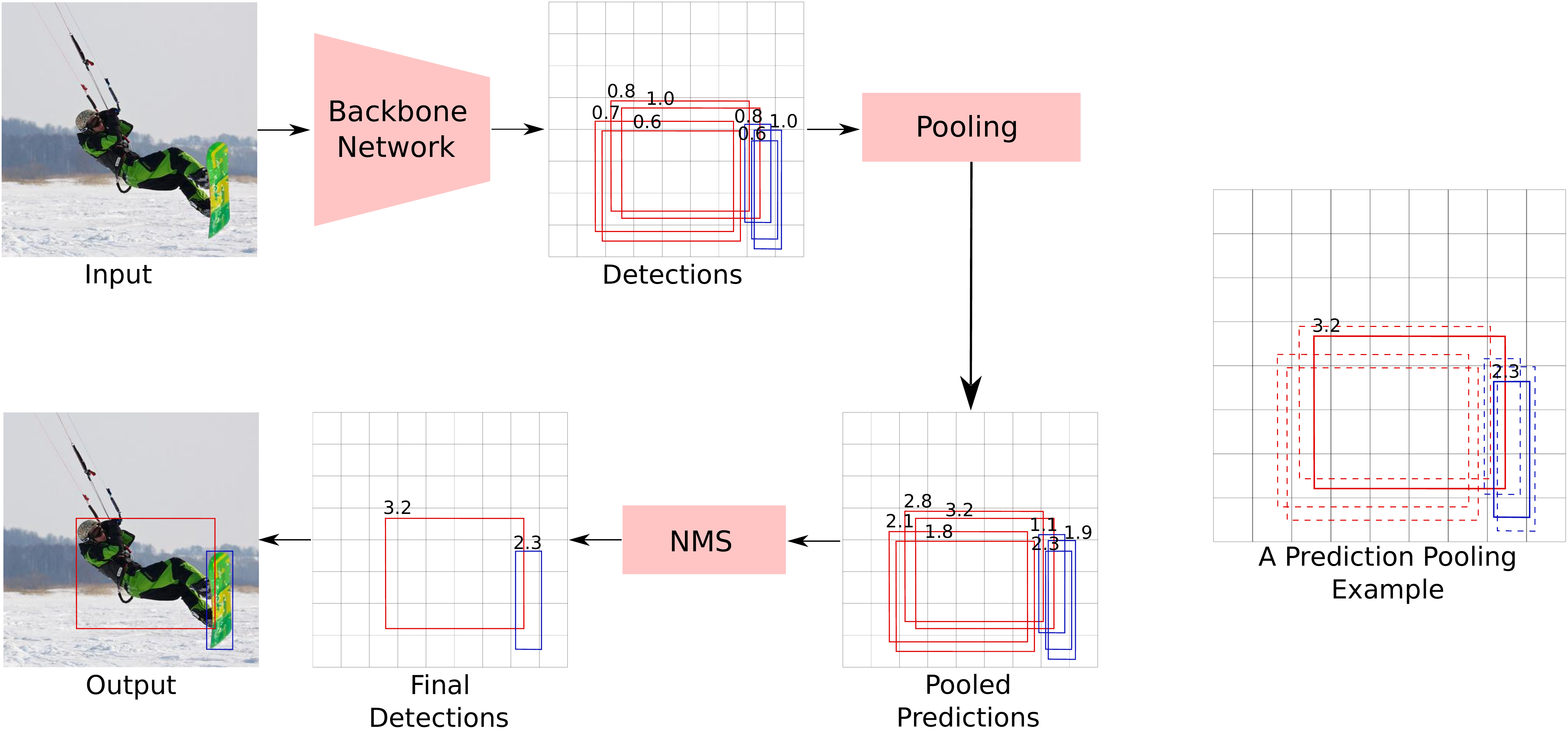}
\caption{(Left) Illustration of PPDet’s inference pipeline. Predicted boxes for \textit{person} and \textit{snowboard} are shown in red and blue, respectively. Red and blue boxes vote for each other among themselves. See text for details. (Right) A pooling example. The dashed-boundary red boxes vote for the solid red box and the dashed-boundary blue boxes vote for the solid blue box. Final scores (after aggregation) of solid boxes are shown.}
\label{fig:inference}
\end{figure}

\paragraph{Inference.}  Inference pipeline of PPDet is given in Figure~\ref{fig:inference}. First, the input image is fed to a backbone neural network model (described in the next section) which produces the initial set of detections. Each detection is associated with (i) a bounding box, (ii) an object class (chosen as the class with maximum probability) and (iii) a confidence score. Within these detections, those labeled with the background class are eliminated. We consider each remaining detection at this stage as a vote for the object that it belongs to, where the box is an hypothesis for the location of the object and the confidence score is the strength of the vote. Next, these detections are pooled together  as follows. If two detections belonging to the same object class overlap more than a certain amount (i.e. intersection over union (IoU) $>0.6$), then we consider them as voting for the same object and the score of each detection is increased by $k^ {(IoU - 1.0)}$ times the score of the other detection, where $k$ is a constant. The more the IoU, the higher the increase. After applying this process to every pair of detections, we obtain the scores for final detections. This step is followed by the class aware non-maxima suppression (NMS) operation which yields the final detections.

Note that although the prediction pooling used in inference might seem to be different from the pooling employed in training, in fact, they are the same process. The pooling used in training makes the assumption that the bounding boxes predicted by the features in the positive area overlap among each other perfectly (i.e. IoU=1).


\paragraph{Network architecture.} PPDet uses the network model of RetinaNet~\cite{retinanet} which consists of a backbone convolutional neural network (CNN) followed by a feature pyramid network (FPN)~\cite{fpn}. The FPN computes a multi-scale feature representation and produces feature maps at five different scales. There are two separate, parallel networks on the top of each FPN layer, namely classification network and regression network. The classification network outputs a $W \times H \times C$ tensor where $W$ and $H$ are spatial dimensions (width and height, respectively) and $C$ is the number of the classes. Similarly, the regression network outputs a $W \times H \times 4$ tensor where $4$ is the number of bounding box coordinates. We refer to each pixel in these tensors as a \emph{feature}.

\section{Experiments}

This section describes the experiments we conducted to show the effectiveness of our proposed method. First, we present ablation  experiments to find the optimal relative area of the positive region  within GT boxes and the regression loss weight. Next, we present several performance comparisons on the COCO dataset. Finally, we provide sample heatmaps which show the GT box relative locations of features responsible for correct detections.

\paragraph{Implementation Details.} We use Feature Pyramid Network (FPN)~\cite{fpn} on top of ResNet~\cite{resnet} and ResNeXt~\cite{resnext} as our backbone networks for ablations and state of the art comparison, respectively. For all experiments, we resize the images such that their shorter side is $800$ pixels and longer side is maximum $1300$ pixels. The constant $k$ used in vote aggregation (i.e., $k^{\mathrm{IoU}-1}$) was set to $40$ experimentally. We trained all of the experiments on 4 Tesla V100 GPUs, and tested using a single Tesla V100 GPU. We used MMDetection~\cite{mmdetection} framework with Pytorch~\cite{pytorch} to implement our models.

\subsection{Ablation Experiments}
Unless stated otherwise, in ablation experiments we used ResNet-50 with FPN backbone. They are trained with a batch size of 16 for 12 epochs using stochastic gradient descent (SGD) with weight decay of $0.0001$ and momentum of $0.9$. Initial learning rate $0.01$ was dropped $10\times$ at epochs 8 and 11. All ablation models are trained on COCO~\cite{mscoco} \texttt{train2017} dataset and tested on \texttt{val2017} set.

\paragraph{Size of the ``positive area’’.} As explained before, we define the ``positive area’’ as the region that is co-centric with the GT box and that has the same shape as the GT box. We adjust the size of this ``positive area’’ by multiplying its width and height with a shrink factor. We experimented with shrink factors between $1.0$ and $0.2$. Performance results are presented in Table~\ref{table:ratio_exps}. From shrink factor $1.0$ to $0.4$, \textit{AP} increases, however, after that point performance degrades  dramatically. Based on this ablation, we set the shrink factor to $0.4$ for the rest of our experiments.

\begin{table}[h]
\centering
\resizebox{0.6\columnwidth}{!}{
  \begin{tabular}{lccccccc}
    \toprule 
    \textbf{Shrink Factor} & \textit{AP}& \textit{AP$_{50}$} &  \textit{AP$_{75}$} & \textit{AP$_{S}$} & \textit{AP$_{M}$} & \textit{AP$_{L}$} \\
    \midrule 
     1.0 & 30.0 & 44.4 & 32.5 & 17.2 & 33.9 & 38.4         \\
     0.8 & 32.4 & 47.9 & 35.1 & 18.0 & 36.4 & 41.9         \\
     0.6 & 34.5 & 51.3 & 37.5 & 19.6 & 38.5 & 44.5         \\
     0.4         & \textbf{36.0} & \textbf{53.6}        & \textbf{39.0}&\textbf{20.4}& \textbf{39.6}& \textbf{46.6} \\
     0.2 & 32.6 & 50.3 & 34.7 & 17.9 & 36.3 & 42.5         \\
   \bottomrule 
  \end{tabular}}
\caption{Experiments to determine the best shrink factor which defines the relative size of the ``positive area’’ with respect to the GT box. Models were trained on \texttt{train2017} and results were obtained  on \texttt{val2017}.}
 \label{table:ratio_exps}
\end{table}

\paragraph{Regression loss weight.} To find the optimal balance between the classification and regression loss, we conducted ablation experiments on the regression loss weight. As shown in  Table~\ref{table:weight_exps}, $0.75$ yields the best results. We set the weight of the regression loss to $0.75$ for the rest of our experiments.

\begin{table}[h]
\centering
\resizebox{0.6\columnwidth}{!}{
  \begin{tabular}{lccccccc}
    \toprule 
    \textbf{RL weight} & \textit{AP}& \textit{AP$_{50}$} &  \textit{AP$_{75}$} & \textit{AP$_{S}$} & \textit{AP$_{M}$} & \textit{AP$_{L}$} \\
    \midrule 
     1.00 & 36.0 & 53.6 & 39.0 & 20.4 & 39.6 & 46.6         \\
     0.90 & 36.0 & 53.9 & 39.3 & 20.1 & 39.6 & 47.2         \\
     0.75 & \textbf{36.3} & 54.3 & \textbf{39.5}& \textbf{21.1}& 39.5 & \textbf{47.5} \\
     0.60 & 36.2 & \textbf{54.6} & \textbf{39.5} & 21.0 & \textbf{40.1} & 47.1         \\
\bottomrule 
  \end{tabular}}
\caption{Experiments on regression loss (RL) weight. Models were trained  on \texttt{train2017} and results were obtained on \texttt{val2017}.}
 \label{table:weight_exps}
\end{table}

\paragraph{Improvements.} We also employed improvements used in other state-of-the-art object detectors~\cite{fcos, foveabox, centernet}. First, we trained our baseline model using ResNet-101 with FPN backbone. Later, we replaced the last convolution layer before class prediction in the classification branch with deformable convolutional layers. This modification improved the performance  around $0.3$ for all \textit{APs} (see Table~\ref{table:improvements}). Later, on top of this modification, we add another one where we adopt group normalization after each convolution layer in the regression and classification branches. As seen in Table~\ref{table:improvements}, this modification increased \textit{AP} by $0.6$ and \textit{AP$_{50}$} by $1.1$. In this table, we also provide results for the recently introduced \textit{moLRP}~\cite{lrp} metric, which combines localization, precision and recall in a single metric. Lower values are better. Models are trained with a batch size of 16 for 24 epochs using stochastic gradient descent (SGD) with weight decay of $0.0001$ and momentum of $0.9$. Initial learning rate $0.01$ was dropped $10\times$ at epochs 16 and 22. We include these two modifications in our final model.

\paragraph{Class imbalance.} PPDet sum-pools predictions into a single prediction per object instance which reduces the number of positives during training. One may think that it exacerbates the class imbalance~\cite{imbalance} even more. To analyse the issue, we calculated the average number of positives per image, which is $7$ for PPDet, $41$ for FoveBox and $165$ for RetinaNet. PPDet considerably decreases the number of positives. However, this is still small compared to the number of negatives (tens of thousands), hence, it does not exacerbate the existing class imbalance problem. We use focal loss to tackle the imbalance.

\begin{table}
\centering
\resizebox{0.60\columnwidth}{!}{
 \begin{tabular}{lccccccc}
    \toprule 
    \textbf{Method} & \textit{AP}& \textit{AP$_{50}$} &  \textit{AP$_{75}$} & \textit{AP$_{S}$} & \textit{AP$_{M}$} & \textit{AP$_{L}$} & \textit{moLRP} $\downarrow$  \\
    \midrule 
     Baseline  & 39.6 & 58.0 & 43.4 & 23.9 & 44.1 & 51.0 & 68.9 \\
     + Deform. Conv. & 39.9 & 58.4  & 43.7 & 24.2 & 44.4 & 51.3 & 68.7 \\
     + Group Norm.         & \textbf{40.5} & \textbf{59.5}        & \textbf{44.2}        & \textbf{25.4}        & 
\textbf{44.7}        & \textbf{52.3}        & \textbf{67.8} \\
   \bottomrule 
  \end{tabular}}
\caption{Experiments on improvements. Using deformable convolution in the classification branch and group normalization layers further improve detection performances. Models are trained on \texttt{train2017} and tested on \texttt{val2017} set.}
 \label{table:improvements}
\end{table}


\begin{table}[H]
\resizebox{0.99\columnwidth}{!}{
 \begin{tabular}{llccccccccc}
   \toprule 
  Method & Backbone & Train size & Test size &  \textit{AP}& \textit{AP$_{50}$} &  \textit{AP$_{75}$} & \textit{AP$_{S}$} & \textit{AP$_{M}$} & \textit{AP$_{L}$} & FPS\\
   \midrule 
   \textbf{\textit{Two-stage detectors:}} & & & & & & & & & & \\
R-FCN~\cite{rfcn} & ResNet-101 &  800$\times$800 & 600$\times$600 & 29.9 & 51.9 &  -  & 10.8 & 32.8 & 45.0 & 5.9 \\
CoupleNet~\cite{couplenet} & ResNet-101 &  ori. & ori. &  34.4 & 54.8 & 37.2 & 13.4 &  38.1 & 50.8 & -\\
Faster R-CNN+++~\cite{resnet} & ResNet-101 &  1000$\times$600 & 1000$\times$600& 34.9 & 55.7& 37.4& 15.6& 38.7 & 50.9 & - \\
Faster R-CNN~\cite{fpn} & ResNet-101-FPN & 1000$\times$600 & 1000$\times$600& 36.2 & 59.1 & 39.0& 18.2 &39.0&48.2 & 5.0 \\
Mask R-CNN~\cite{mask} & ResNeXt-101-FPN &   1300$\times$800 & 1300$\times$800& 39.8 &  62.3 & 43.4 & 22.1 & 43.2 & 51.2 & 11.0 \\
Cascade R-CNN~\cite{cascade} & ResNet-101 &   - & - & 42.8 & 62.1& 46.3 & 23.7& 45.5 &55.2& \textbf{12.0} \\
PANet~\cite{panet} & ResNeXt-101 & 1400$\times$840 & 1400$\times$840& \textbf{47.4} & \textbf{67.2} & \textbf{51.8} & \textbf{30.1} & \textbf{51.7} & \textbf{60.0} & - \\
   \midrule 
\textbf{\textit{One-stage, anchor-based:}} & & & & & & & & & &\\

SSD~\cite{ssd}         & VGG-16                 &  512$\times$512 & 512$\times$512 & 28.8 & 48.5 & 30.3 & 10.9 & 31.8 & 43.5         & -\\
YOLOv3~\cite{yolo3}        & Darknet-53                  &  608$\times$608 & 608$\times$608& 33.0 & 57.9 &  34.4 & 18.3 &  35.4 & 41.9 & \textbf{20.0}\\
DSSD513~\cite{dssd} & ResNet-101         &   513$\times$513 & 513$\times$513& 33.2& 53.3 & 35.2 & 13.0 & 35.4 &51.1        & - \\ 
RefineDet (SS)~\cite{refinedet} & ResNet-101 &  512$\times$512 & 512$\times$512&  36.4 & 57.5 & 39.5 & 16.6 & 39.9 & 51.4 & - \\
RetinaNet~\cite{retinanet} & ResNet-101-FPN &  1300$\times$800 & 1300$\times$800 & 39.1 & 59.1 & 42.3 & 21.8 & 42.7 & 50.2 & 10.9$^*$ \\
RetinaNet~\cite{retinanet} & ResNeXt-101-FPN &  1300$\times$800 & 1300$\times$800 & 40.8 & 61.1 & 44.1 & 24.1 & 44.2 & 51.2 & 7.0$^*$ \\
RefineDet (MS)~\cite{refinedet} & ResNet-101 &  512$\times$512 & $\leq$2.25$\times$ & 41.8 & 62.9 & 45.7 &  25.6 & 45.1 & 54.1 & - \\
GA-RetinaNet~\cite{guided}$^*$& ResNet-101 &  1300$\times$960 & 1300$\times$800 & 41.9 & 62.2 & 45.3 & 24.0 & 45.3 & 53.8 & - \\
FreeAnchor (SS)~\cite{freeanchor} & ResNeXt-101-FPN &   1300$\times$960 & 1300$\times$960 & 44.9 & 64.3 & 48.5 & 26.8 & 48.3 & 55.9 & 8.4$^*$ \\
FreeAnchor (MS)~\cite{freeanchor} & ResNeXt-101-FPN &  1300$\times$960 & $\sim\leq$2.0$\times$ & \textbf{47.3} &   \textbf{66.3} &  \textbf{51.5} &   \textbf{30.6} & \textbf{50.4} &  \textbf{59.0} & - \\
\midrule 


\textbf{\textit{Anchor-free, bottom-up:}} & & & & & & & & & & \\

ExtremeNet (SS)~\cite{extremenet} & Hourglass-104 &   511$\times$511 & ori. & 40.2 & 55.5 & 43.2 & 20.4 & 43.2 & 53.1  & 3.1 \\
CornerNet (SS)~\cite{cornernet} & Hourglass-104 & 511$\times$511 & ori.& 40.5 & 56.5 & 43.1 & 19.4 & 42.7 & 53.9 & 4.1 \\
CornerNet  (MS)~\cite{cornernet}& Hourglass-104 &  511$\times$511 & $\leq$1.5$\times$ & 42.1 & 57.8 & 45.3 & 20.8 & 44.8 & 56.7 & - \\
CenterNet (SS)~\cite{centernet} & Hourglass-104 &   512$\times$512 & ori. & 42.1 & 61.1 & 45.9 &  24.1 & 45.5 & 52.8 & \textbf{7.8} \\

HoughNet (SS)~\cite{houghnet} & Hourglass-104 &   512$\times$512 & $\leq$ ori.$\times$ & 43.1 & 62.2 & 46.8 &  24.6 & 47.0 & 54.4  & 6.4 \\

ExtremeNet (MS)~\cite{extremenet} & Hourglass-104 &  511$\times$511 & $\leq$1.5$\times$ & 43.7 & 60.5 & 47.0 & 24.1 & 46.9 & 57.6  & - \\
CenterNet (SS)~\cite{triplet} & Hourglass-104 &  511$\times$511 & ori. & 44.9 & 62.4 & 48.1 & 25.6 &  47.4 & 57.4 & 3.0  \\
CenterNet (MS)~\cite{centernet} & Hourglass-104 &   512$\times$512 & $\leq$1.5$\times$ & 45.1 & 63.9 & 49.3 &  26.6 & 47.1 & 57.7  & - \\
HoughNet (MS)~\cite{houghnet} & Hourglass-104 &   512$\times$512 & $\leq$1.8$\times$ & 46.4 & \textbf{65.1} & \textbf{50.7} &  \textbf{29.1} & 48.5 & 58.1  & - \\
CenterNet (MS)~\cite{triplet} & Hourglass-104 & 511$\times$511 & $\leq$1.8$\times$ &  \textbf{47.0} &  64.5 &  \textbf{50.7} & 28.9 &   \textbf{49.9} &  \textbf{58.9}  & - \\

\textbf{\textit{Anchor-free, top-down:}} & & & & & & & & & & \\
FoveaBox~\cite{foveabox} (SS)  &  ResNet-101-FPN & 1300$\times$800 & 1300$\times$800 & 40.6 & 60.1 & 43.5 & 23.3 & 45.2 & 54.5 & - \\
FoveaBox~\cite{foveabox} (SS) &  ResNeXt-101-FPN & 1300$\times$800 & 1300$\times$800 & 42.1 & 61.9 & 45.2 &  24.9 &  46.8 &  55.6 & - \\
FSAF (SS)~\cite{fsaf} & ResNeXt-101-FPN &  1300$\times$800 & 1300$\times$800 &   42.9  &  63.8 &  46.3 &  26.6 &  46.2  &  52.7 &  2.7 \\
FoveaBox~\cite{foveabox} (MS)$^\dagger$ &  ResNet-101-FPN & 1300$\times$800 & 1300$\times$800 & 44.2 & \textbf{65.4} & 47.8 &  28.8 &  46.7 &  53.7 & - \\
FSAF (MS)~\cite{fsaf} &ResNeXt-101-FPN &   1300$\times$800 & $\sim\leq$2.0$\times$ &   44.6 &  65.2 &  48.6 &  29.7 &  47.1  &  54.6 &  - \\
FCOS~\cite{fcos}  & ResNeXt-101-FPN  &   1300$\times$800 & 1300$\times$800 & 44.7 & 64.1 &  48.4 &  27.6 &  47.5 &    55.6 & 7.0$^*$  \\

\midrule 
PPDet (SS) & ResNet-101-FPN & 1300$\times$800 & 1300$\times$800 &  40.7 &   60.2 &   44.5 & 24.5 &  44.4 & 49.7 & \textbf{7.5}  \\
PPDet (SS) & ResNeXt-101-FPN & 1300$\times$800 & 1300$\times$800 &  42.3 &   62.0 &   46.3 & 26.2 &  46.0 & 51.9 & 4.1 \\
PPDet (MS) & ResNet-101-FPN & 1300$\times$800 & $\sim\leq$2.0$\times$ & 45.2 & 63.5 &   50.3 & 30.0 &  48.6 & 54.7 & - \\
PPDet (MS) & ResNeXt-101-FPN & 1300$\times$800 & $\sim\leq$2.0$\times$ & \textbf{46.3} &  64.8 &  \textbf{51.6} & \textbf{31.4} &  \textbf{49.9} & \textbf{56.4} & - \\
  \bottomrule 
\end{tabular}}
\caption{Detection performances on COCO \texttt{test-dev} set. The methods are divided into three groups: two-stage, one-stage anchor-based and one-stage anchor-free. The best results are boldfaced separately for each group. PPDet achieves state-of-the-art results on the ${AP}_{S}$ metric among all the detectors. $^*$ results are taken from MMDetection. $^\dagger$ MS test for FoveaBox is implemented by us on top of the original code.}
\label{table:stateoftheartcomp}
\end{table}

\subsection{State-of-the-art comparison}
To compare our model with the state-of-the-art methods, we used ResNet-101 with FPN and ResNeXt-101-64x4d with FPN backbones. They are trained with batch sizes of 16 and 8 for 24 and 16 epochs, respectively, using SGD with weight decay of $0.0001$ and momentum of $0.9$. For the ResNet backbone, initial learning rate $0.01$ was dropped $10\times$ at epochs 16 and 22. For the ResNeXt backbone, initial learning rate $0.005$ was dropped $10\times$ at epochs 11 and 14. The models are trained on COCO~\cite{mscoco} \texttt{train2017} dataset and tested on \texttt{test-dev} set. We used $(800, 480)$, $(1067, 640)$, $(1333, 800)$, $(1600, 960)$, $(1867, 1120)$, $(2133, 1280)$  scales for multi-scale testing. 
Table~\ref{table:stateoftheartcomp} presents performances of PPDet and several established state-of-the-art detectors.

FSAF~\cite{fsaf} and FoveaBox~\cite{foveabox} use a similar approach to ours to build the ``positive area’’. While single scale testing performance of PPDet  is  comparable with that of FSAF on the same  ResNeXt-101-64x4d with FPN backbone,  PPDet’s multi-scale testing performance is $1.7$ \textit{AP} points better than that of FSAF’s. Our both models with single-scale testing get slightly better results than FoveaBox while outperforming it on small objects by more than $1.0$.  The results of our multi-scale testing outperforms FoveaBox by $1$ \textit{AP} on the same ResNet-101 with FPN backbone.

Our multi-scale performance is the best among all the anchor-free top-down methods. Moreover, our multi-scale performance  on small objects (i.e. \textit{AP$_{S}$}) sets the new  state-of-the-art among all detectors in Table~\ref{table:stateoftheartcomp}.

We conducted experiments to analyse the effect of the prediction pooling for training and inference. When we removed the prediction pooling from the inference pipeline of our ResNet-101-FPN backbone model, we observed that \textit{AP} goes down by $2.5$ points on \texttt{val2017} set. To analyse the effect of prediction pooling for training, we added prediction pooling to RetinaNet~\cite{retinanet} and FoveaBox~\cite{foveabox} only during inference (so, no PP in training). This resulted in $0.5$ and $2.8$ points drop in  \textit{AP} for RetinaNet and FoveaBox, respectively.

We also conducted another experiment to test the effectiveness of sum-pooling over max-pooling. For max-pooling, we identified the feature within the positive area, whose predicted box overlaps the most with the GT box. Then, only this feature is included in focal loss to represent its GT box during training.  This strategy dropped \textit{AP} by more than $2$ points, yielding $38.4$ with ResNet101 with FPN backbone.

As an additional result, we present the performance of PPDet on the PASCAL VOC dataset~\cite{pascal}. For training, we used the union set of PASCAL VOC 2007 \texttt{trainval} and VOC 2012 \texttt{trainval} images (``07+12’’). For testing, we used the \texttt{test} set of PASCAL VOC 2007. Our PPDet model achieves $77.8$ mean average precision (mAP) outperforming FoveaBox~\cite{foveabox} at $76.6$ mAP, which we consider as a baseline here, when both use the ResNet-50 backbone.

\begin{figure}[h]
\centering
\begin{tabular}{ccccccr}

 Person &  Bicycle & Boat & Bench & Tie & Skis & \\
\includegraphics[width=0.14\textwidth,  ,valign=m, keepaspectratio,]{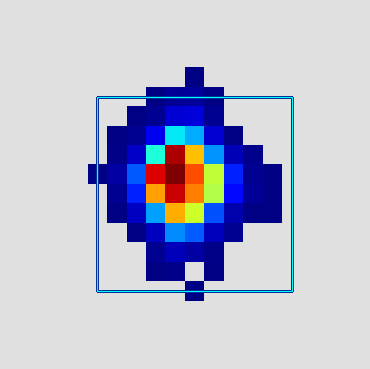}   \hspace*{-12pt}& 
\includegraphics[width=0.14\textwidth,  ,valign=m, keepaspectratio,]{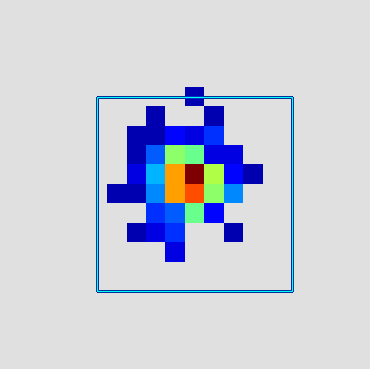} \hspace*{-12pt} &
\includegraphics[width=0.14\textwidth,  ,valign=m, keepaspectratio,]{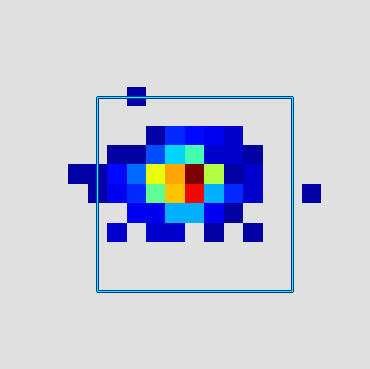}   \hspace*{-12pt}&
\includegraphics[width=0.14\textwidth,   ,valign=m, keepaspectratio,]{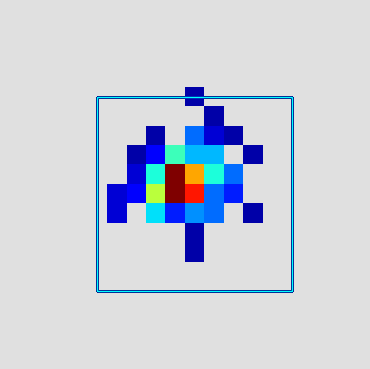}   \hspace*{-12pt}&
\includegraphics[width=0.14\textwidth,   ,valign=m, keepaspectratio,]{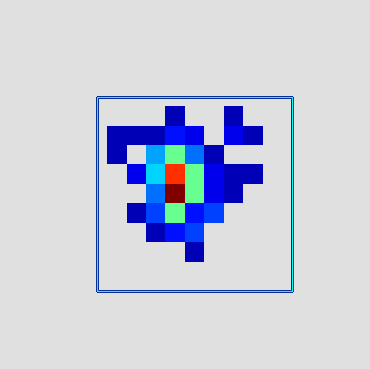}   \hspace*{-12pt}&
\includegraphics[width=0.14\textwidth,   ,valign=m, 
keepaspectratio,]{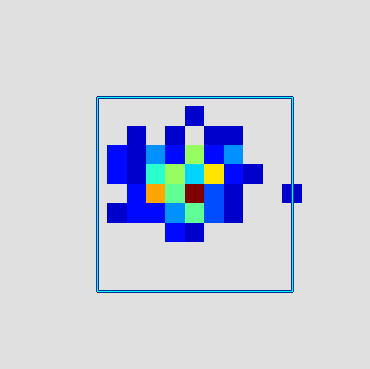}  \hspace*{-12pt} &
 \\
\cmidrule(lr){1-6}
\includegraphics[width=0.14\textwidth,  ,valign=m, 
keepaspectratio,]{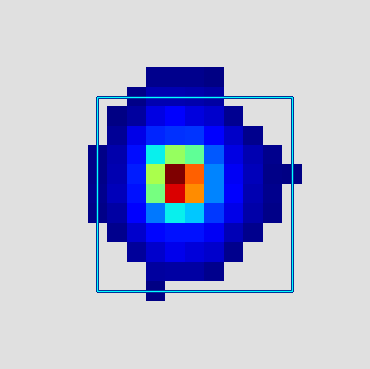}   \hspace*{-12pt} & 
\includegraphics[width=0.14\textwidth,  ,valign=m, keepaspectratio,]{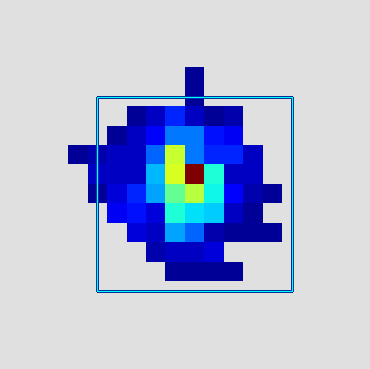}  \hspace*{-12pt}&
\includegraphics[width=0.14\textwidth,  ,valign=m, 
keepaspectratio,]{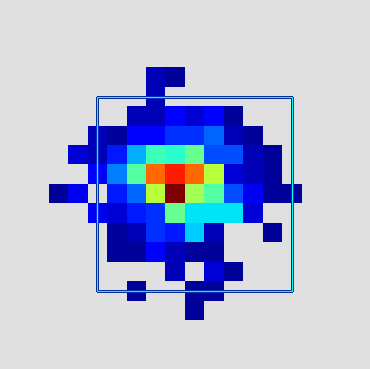}  \hspace*{-12pt} &
\includegraphics[width=0.14\textwidth,   ,valign=m, keepaspectratio,]{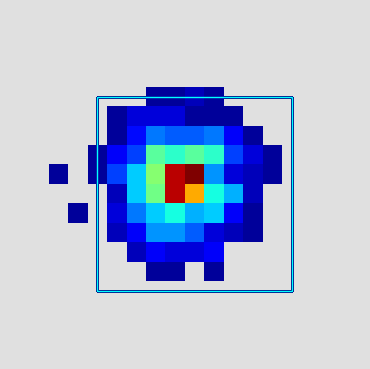}   \hspace*{-12pt} &
\includegraphics[width=0.14\textwidth,   ,valign=m, keepaspectratio,]{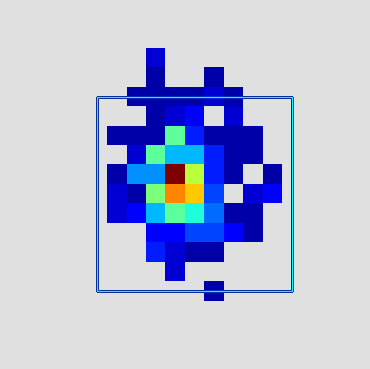}  \hspace*{-12pt} &
\includegraphics[width=0.14\textwidth,   ,valign=m, 
keepaspectratio,]{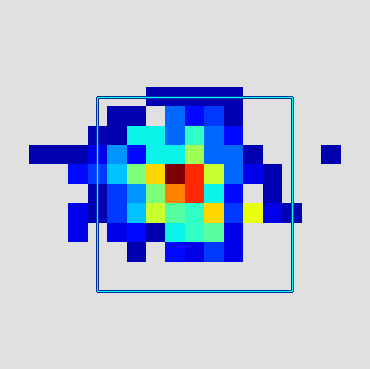}  \hspace*{-12pt} &
\includegraphics[width=0.045\textwidth,   ,valign=m, 
keepaspectratio,]{./figures/heatmaps/colormap.png}   \hspace*{-12pt}  \\

\cmidrule(lr){1-6}
\includegraphics[width=0.14\textwidth,  ,valign=m, 
keepaspectratio,]{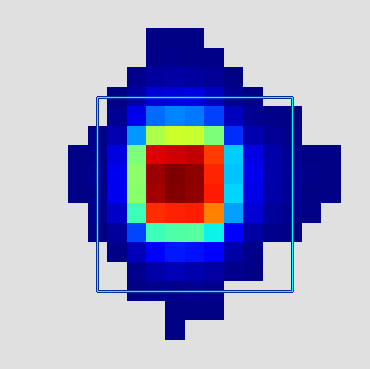}   \hspace*{-12pt} & 
\includegraphics[width=0.14\textwidth,  ,valign=m, keepaspectratio,]{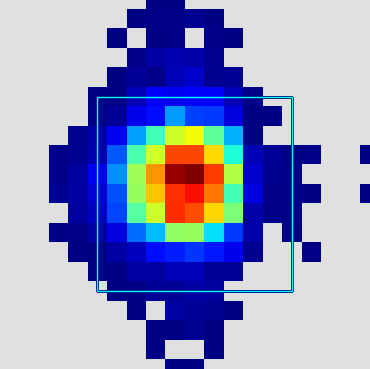}  \hspace*{-12pt}&
\includegraphics[width=0.14\textwidth,  ,valign=m, 
keepaspectratio,]{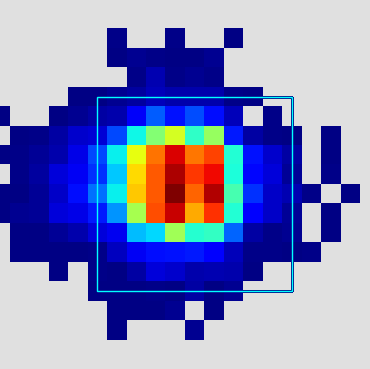}  \hspace*{-12pt} &
\includegraphics[width=0.14\textwidth,   ,valign=m, keepaspectratio,]{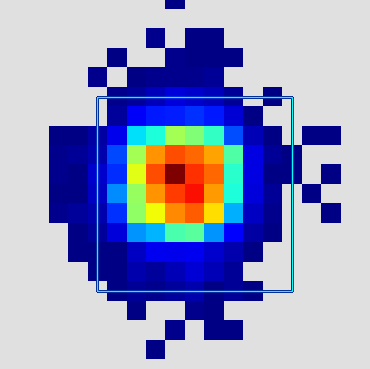}   \hspace*{-12pt} &
\includegraphics[width=0.14\textwidth,   ,valign=m, keepaspectratio,]{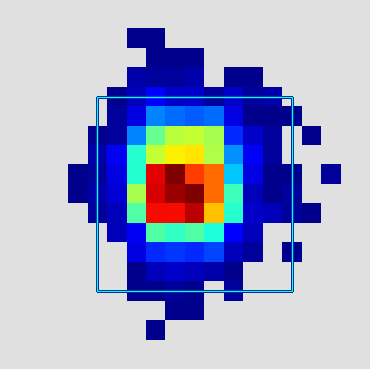}  \hspace*{-12pt} &
\includegraphics[width=0.14\textwidth,   ,valign=m, 
keepaspectratio,]{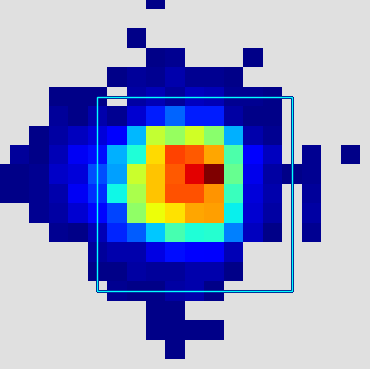}  \hspace*{-12pt} &  \\
\end{tabular}
\caption{
Feature locations that are responsible for detection during inference, relative to the ground-truth box (blue rectangle). To bring different ground-truth boxes into the same plot, we normalized each ground-truth box to a canonical size. Relative locations of responsible features were normalized accordingly. {\bf Top row} shows the heatmap of responsible feature locations for anchor-based RetinaNet. {\bf Second row} shows the same for anchor-free object detector FoveaBox. {\bf Bottom row} shows the same for PPDet. Heatmaps were obtained on COCO \texttt{val2017} images with ResNet-101 with FPN backbone. RetinaNet detects objects mostly with center cells. In terms of peakyness, FoveaBox’s heatmaps are similar to RetinaNet’s. PPDet detects objects from a wider area, also from outside of the object box. 
}
\label{fig:heatmaps}
\end{figure}

Figure~\ref{fig:heatmaps} shows the heatmap of cell centers relative to the ground-truth box, which are responsible for detection. The heatmaps of RetinaNet are concentrated at the center of the ground-truth object boxes. In contrast, PPDet’s final detections are formed from a relatively wider area verifying its dynamic and automatic characteristics on assigning
weights to the features in the positive area. In addition to the detections coming from the center of the ground-truth box, they may heavily come from the different parts of the ground-truth box.

\arrayrulecolor{white}

\section{Conclusion}
In this work, we introduced a novel labeling strategy for the training of anchor-free object detectors. While current anchor-free methods force positive labels on all the features that are  spatially inside a predefined central region of a ground-truth box, our labeling strategy relaxes this constraint by sum-pooling predictions stemming from individual features into a single prediction. This allows the model to reduce the contributions of non-discriminatory features during training. We developed PPDet, a one-stage, anchor-free object detector which employs the new labeling strategy during training and a new inference method based on pooling predictions. We analyzed our idea by conducting several ablation experiments. We reported results on COCO \texttt{test-dev} and show that PPDet performs on par with the state-of-the-art and achieves state-of-the-art results on small objects (\textit{AP$_{S}$} $31.4$). We further validated the effectiveness of our method through visual inspections.  
 


\paragraph*{\textbf{Acknowledgments}}
This work was supported by the Scientific and Technological Research Council of Turkey (T\"{U}B\.{I}TAK) through the project titled ``Object Detection in Videos with Deep Neural Networks" (grant \#117E054). The numerical calculations reported in this paper were partially performed at T\"{U}B\.{I}TAK ULAKB\.{I}M, High Performance and Grid Computing Center (TRUBA resources). We also gratefully acknowledge the support of the AWS Cloud Credits for Research program.



{ \small
\bibliographystyle{ieeetran}
\bibliography{references} }

\begin{thebibliography}{34}
\providecommand{\natexlab}[1]{#1}
\providecommand{\url}[1]{\texttt{#1}}
\expandafter\ifx\csname urlstyle\endcsname\relax
  \providecommand{\doi}[1]{doi: #1}\else
  \providecommand{\doi}{doi: \begingroup \urlstyle{rm}\Url}\fi

\bibitem[Cai and Vasconcelos(2018)]{cascade}
Zhaowei Cai and Nuno Vasconcelos.
\newblock Cascade {R-CNN}: Delving into high quality object detection.
\newblock In \emph{{IEEE} Conference on Computer Vision and Pattern
  Recognition}, pages 6154--6162, 2018.

\bibitem[Chen et~al.(2019)Chen, Wang, Pang, Cao, Xiong, Li, Sun, Feng, Liu, Xu,
  Zhang, Cheng, Zhu, Cheng, Zhao, Li, Lu, Zhu, Wu, Dai, Wang, Shi, Ouyang, Loy,
  and Lin]{mmdetection}
Kai Chen, Jiaqi Wang, Jiangmiao Pang, Yuhang Cao, Yu~Xiong, Xiaoxiao Li,
  Shuyang Sun, Wansen Feng, Ziwei Liu, Jiarui Xu, Zheng Zhang, Dazhi Cheng,
  Chenchen Zhu, Tianheng Cheng, Qijie Zhao, Buyu Li, Xin Lu, Rui Zhu, Yue Wu,
  Jifeng Dai, Jingdong Wang, Jianping Shi, Wanli Ouyang, Chen~Change Loy, and
  Dahua Lin.
\newblock {MMDetection}: Open mmlab detection toolbox and benchmark.
\newblock \emph{arXiv preprint arXiv:1906.07155}, 2019.

\bibitem[Dai et~al.(2016)Dai, Li, He, and Sun]{rfcn}
Jifeng Dai, Yi~Li, Kaiming He, and Jian Sun.
\newblock R-fcn: Object detection via region-based fully convolutional
  networks.
\newblock In \emph{Advances in neural information processing systems}, pages
  379--387, 2016.

\bibitem[Duan et~al.(2019)Duan, Bai, Xie, Qi, Huang, and Tian]{triplet}
Kaiwen Duan, Song Bai, Lingxi Xie, Honggang Qi, Qingming Huang, and Qi~Tian.
\newblock Centernet: Keypoint triplets for object detection.
\newblock In \emph{{IEEE} International Conference on Computer Vision}, 2019.

\bibitem[Everingham et~al.(2010)Everingham, Van~Gool, Williams, Winn, and
  Zisserman]{pascal}
M.~Everingham, L.~Van~Gool, C.~K.~I. Williams, J.~Winn, and A.~Zisserman.
\newblock The {PASCAL} {V}isual {O}bject {C}lasses {(VOC)} {C}hallenge.
\newblock \emph{International Journal of Computer Vision}, pages 303--338,
  2010.

\bibitem[Fu et~al.(2017)Fu, Liu, Ranga, Tyagi, and Berg]{dssd}
Cheng-Yang Fu, Wei Liu, Ananth Ranga, Ambrish Tyagi, and Alexander~C Berg.
\newblock Dssd: Deconvolutional single shot detector.
\newblock \emph{arXiv preprint arXiv:1701.06659}, 2017.

\bibitem[Girshick(2015)]{fastrcnn}
Ross Girshick.
\newblock Fast {R-CNN}.
\newblock In \emph{{IEEE} International Conference on Computer Vision}, 2015.

\bibitem[He et~al.()He, Gkioxari, Doll{\'a}r, and Girshick]{mask}
Kaiming He, Georgia Gkioxari, Piotr Doll{\'a}r, and Ross~B. Girshick.
\newblock Mask r-cnn.
\newblock \emph{{IEEE} International Conference on Computer Vision}, pages
  2980--2988.

\bibitem[He et~al.(2016)He, Zhang, Ren, and Sun]{resnet}
Kaiming He, Xiangyu Zhang, Shaoqing Ren, and Jian Sun.
\newblock Deep residual learning for image recognition.
\newblock In \emph{{IEEE} Conference on Computer Vision and Pattern
  Recognition}, pages 770--778, 2016.

\bibitem[Kong et~al.(2020)Kong, Sun, Liu, Jiang, Li, and Shi]{foveabox}
Tao Kong, Fuchun Sun, Huaping Liu, Yuning Jiang, Lei Li, and Jianbo Shi.
\newblock Foveabox: Beyond anchor-based object detector.
\newblock \emph{IEEE Transactions on Image Processing}, pages 7389--7398, 2020.

\bibitem[Law and Deng(2018)]{cornernet}
Hei Law and Jia Deng.
\newblock Cornernet: Detecting objects as paired keypoints.
\newblock In \emph{{IEEE} European Conference on Computer Vision}, pages
  734--750, 2018.

\bibitem[Lin et~al.(2014)Lin, Maire, Belongie, Hays, Perona, Ramanan,
  Doll{\'a}r, and Zitnick]{mscoco}
Tsung-Yi Lin, Michael Maire, Serge Belongie, James Hays, Pietro Perona, Deva
  Ramanan, Piotr Doll{\'a}r, and C~Lawrence Zitnick.
\newblock Microsoft {COCO}: {C}ommon objects in context.
\newblock In \emph{{IEEE} European Conference on Computer Vision}, pages
  740--755. Springer, 2014.

\bibitem[Lin et~al.(2017{\natexlab{a}})Lin, Doll{\'{a}}r, Girshick, He,
  Hariharan, and Belongie]{fpn}
Tsung{-}Yi Lin, Piotr Doll{\'{a}}r, Ross~B. Girshick, Kaiming He, Bharath
  Hariharan, and Serge~J. Belongie.
\newblock Feature pyramid networks for object detection.
\newblock In \emph{{IEEE} Conference on Computer Vision and Pattern
  Recognition}, pages 936--944, 2017{\natexlab{a}}.

\bibitem[Lin et~al.(2017{\natexlab{b}})Lin, Goyal, Girshick, He, and
  Doll{\'{a}}r]{retinanet}
Tsung{-}Yi Lin, Priya Goyal, Ross~B. Girshick, Kaiming He, and Piotr
  Doll{\'{a}}r.
\newblock Focal loss for dense object detection.
\newblock In \emph{{IEEE} International Conference on Computer Vision},
  2017{\natexlab{b}}.

\bibitem[Liu et~al.(2018)Liu, Qi, Qin, Shi, and Jia]{panet}
Shu Liu, Lu~Qi, Haifang Qin, Jianping Shi, and Jiaya Jia.
\newblock Path aggregation network for instance segmentation.
\newblock In \emph{{IEEE} Conference on Computer Vision and Pattern
  Recognition}, pages 8759--8768, 2018.

\bibitem[Liu et~al.(2016)Liu, Anguelov, Erhan, Szegedy, Reed, Fu, and
  Berg]{ssd}
Wei Liu, Dragomir Anguelov, Dumitru Erhan, Christian Szegedy, Scott Reed,
  Cheng-Yang Fu, and Alexander~C Berg.
\newblock Ssd: Single shot multibox detector.
\newblock In \emph{{IEEE} European Conference on Computer Vision}, 2016.

\bibitem[Maninis et~al.(2018)Maninis, Caelles, Pont-Tuset, and {Van
  Gool}]{extremenet}
K.K. Maninis, S.~Caelles, J.~Pont-Tuset, and L.~{Van Gool}.
\newblock Deep extreme cut: From extreme points to object segmentation.
\newblock In \emph{{IEEE} Conference on Computer Vision and Pattern
  Recognition}, 2018.

\bibitem[Oksuz et~al.(2018)Oksuz, Cam, Akbas, and Kalkan]{lrp}
Kemal Oksuz, Baris Cam, Emre Akbas, and Sinan Kalkan.
\newblock Localization recall precision ({LRP}): A new performance metric for
  object detection.
\newblock In \emph{{IEEE} European Conference on Computer Vision}, 2018.

\bibitem[Oksuz et~al.(2020)Oksuz, Cam, Kalkan, and Akbas]{imbalance}
Kemal Oksuz, Baris~Can Cam, Sinan Kalkan, and Emre Akbas.
\newblock {Imbalance Problems in Object Detection: A Review}.
\newblock \emph{{IEEE} Transactions on Pattern Analysis and Machine
  Intelligence}, pages 1--1, 2020.

\bibitem[Paszke et~al.(2019)Paszke, Gross, Massa, Lerer, Bradbury, Chanan,
  Killeen, Lin, Gimelshein, Antiga, Desmaison, Kopf, Yang, DeVito, Raison,
  Tejani, Chilamkurthy, Steiner, Fang, Bai, and Chintala]{pytorch}
Adam Paszke, Sam Gross, Francisco Massa, Adam Lerer, James Bradbury, Gregory
  Chanan, Trevor Killeen, Zeming Lin, Natalia Gimelshein, Luca Antiga, Alban
  Desmaison, Andreas Kopf, Edward Yang, Zachary DeVito, Martin Raison, Alykhan
  Tejani, Sasank Chilamkurthy, Benoit Steiner, Lu~Fang, Junjie Bai, and Soumith
  Chintala.
\newblock Pytorch: An imperative style, high-performance deep learning library.
\newblock In \emph{Advances in Neural Information Processing Systems}, pages
  8024--8035. Curran Associates, Inc., 2019.

\bibitem[Redmon and Farhadi(2018)]{yolo3}
Joseph Redmon and Ali Farhadi.
\newblock Yolov3: An incremental improvement.
\newblock \emph{arXiv preprint arXiv:1804.02767}, 2018.

\bibitem[Ren et~al.(2015)Ren, He, Girshick, and Sun]{faster}
Shaoqing Ren, Kaiming He, Ross Girshick, and Jian Sun.
\newblock Faster {R-CNN}: Towards real-time object detection with region
  proposal networks.
\newblock In \emph{Advances in Neural Information Processing Systems}, pages
  91--99, 2015.

\bibitem[Samet et~al.(2020, in press)Samet, Hicsonmez, and Akbas]{houghnet}
Nermin Samet, Samet Hicsonmez, and Emre Akbas.
\newblock Hough{N}et: Integrating near and long-range evidence for bottom-up
  object detection.
\newblock In \emph{{IEEE} European Conference on Computer Vision}, 2020, in
  press.

\bibitem[Shrivastava et~al.(2016)Shrivastava, Gupta, and Girshick]{ohem}
Abhinav Shrivastava, Abhinav Gupta, and Ross Girshick.
\newblock Training region-based object detectors with online hard example
  mining.
\newblock In \emph{{IEEE} Conference on Computer Vision and Pattern
  Recognition}, pages 761--769, 2016.

\bibitem[Tian et~al.(2019)Tian, Shen, Chen, and He]{fcos}
Zhi Tian, Chunhua Shen, Hao Chen, and Tong He.
\newblock Fcos: Fully convolutional one-stage object detection.
\newblock In \emph{{IEEE} International Conference on Computer Vision}, 2019.

\bibitem[Wang et~al.(2019{\natexlab{a}})Wang, Chen, Yang, Loy, and Lin]{guided}
Jiaqi Wang, Kai Chen, Shuo Yang, Chen~Change Loy, and Dahua Lin.
\newblock Region proposal by guided anchoring.
\newblock In \emph{Proceedings of the IEEE Conference on Computer Vision and
  Pattern Recognition}, pages 2965--2974, 2019{\natexlab{a}}.

\bibitem[Wang et~al.(2019{\natexlab{b}})Wang, Chen, Yang, Loy, and
  Lin]{region_proposals}
Jiaqi Wang, Kai Chen, Shuo Yang, Chen~Change Loy, and Dahua Lin.
\newblock Region proposal by guided anchoring.
\newblock In \emph{{IEEE} Conference on Computer Vision and Pattern
  Recognition}, 2019{\natexlab{b}}.

\bibitem[Xie et~al.(2017)Xie, Girshick, Doll{\'a}r, Tu, and He]{resnext}
Saining Xie, Ross Girshick, Piotr Doll{\'a}r, Zhuowen Tu, and Kaiming He.
\newblock Aggregated residual transformations for deep neural networks.
\newblock In \emph{{IEEE} Conference on Computer Vision and Pattern
  Recognition}, pages 1492--1500, 2017.

\bibitem[Yang et~al.(2018)Yang, Zhang, Zhang, and Sun]{metanachor}
Tong Yang, Xiangyu Zhang, Wenqiang Zhang, and Jian Sun.
\newblock Metaanchor: Learning to detect objects with customized anchors.
\newblock In \emph{NIPS}, 2018.

\bibitem[Zhang et~al.(2018)Zhang, Wen, Bian, Lei, and Li]{refinedet}
Shifeng Zhang, Longyin Wen, Xiao Bian, Zhen Lei, and Stan~Z Li.
\newblock Single-shot refinement neural network for object detection.
\newblock In \emph{{IEEE} Conference on Computer Vision and Pattern
  Recognition}, pages 4203--4212, 2018.

\bibitem[Zhang et~al.(2019)Zhang, Wan, Liu, Ji, and Ye]{freeanchor}
Xiaosong Zhang, Fang Wan, Chang Liu, Rongrong Ji, and Qixiang Ye.
\newblock Freeanchor: Learning to match anchors for visual object detection.
\newblock In \emph{Advances in Neural Information Processing Systems}, 2019.

\bibitem[Zhou et~al.(2019)Zhou, Wang, and Kr{\"a}henb{\"u}hl]{centernet}
Xingyi Zhou, Dequan Wang, and Philipp Kr{\"a}henb{\"u}hl.
\newblock Objects as points.
\newblock In \emph{arXiv preprint arXiv:1904.07850}, 2019.

\bibitem[Zhu et~al.(2019)Zhu, He, and Savvides]{fsaf}
Chenchen Zhu, Yihui He, and Marios Savvides.
\newblock Feature selective anchor-free module for single-shot object
  detection.
\newblock In \emph{{IEEE} Conference on Computer Vision and Pattern
  Recognition}, 2019.

\bibitem[Zhu et~al.(2017)Zhu, Zhao, Wang, Zhao, Wu, and Lu]{couplenet}
Yousong Zhu, Chaoyang Zhao, Jinqiao Wang, Xu~Zhao, Yi~Wu, and Hanqing Lu.
\newblock Couplenet: Coupling global structure with local parts for object
  detection.
\newblock In \emph{Proceedings of the IEEE International Conference on Computer
  Vision}, pages 4126--4134, 2017.

\end{thebibliography}
\end{document}